\documentclass[11pt]{article}
\usepackage{xcolor}
\usepackage{booktabs}

\usepackage[final]{acl} 

\usepackage{times}
\usepackage{latexsym}

\usepackage[T1]{fontenc}

\usepackage[utf8]{inputenc}

\usepackage{hyperref}

\usepackage{natbib}

\usepackage{float}

\usepackage{graphicx}
\usepackage{afterpage}

\usepackage{csquotes}

\title{IndicMMLU-Pro: Benchmarking Indic Large Language Models on Multi-Task Language Understanding}

\author{
  \textbf{Sankalp KJ\textsuperscript{1}},
  \textbf{Ashutosh Kumar\textsuperscript{2}},
  \textbf{Laxmaan Balaji\textsuperscript{3}},\\
  \textbf{Nikunj Kotecha\textsuperscript{3}},
  \textbf{Vinija Jain\textsuperscript{4}\thanks{Work done outside position at Meta}},
  \textbf{Aman Chadha\textsuperscript{5}\thanks{Work done outside position at Amazon Gen AI.}},
  \textbf{Sreyoshi Bhaduri\textsuperscript{6}\thanks{Work done outside position at Amazon.}}
  \\
  \textsuperscript{1}Artificial Intelligence Institute, University of South Carolina \\
  \textsuperscript{2}Rochester Institute of Technology
  \textsuperscript{3}Independent Researcher
  \\
  \textsuperscript{4}Meta AI
  \textsuperscript{5}Amazon Gen AI
  \textsuperscript{6}Amazon
\\
  \small sjajee@email.sc.edu, ak1825@rit.edu, laxmaanb@gmail.com, \\ \small kotecha.nikunj95@gmail.com, hi@vinija.ai, hi@aman.ai, sreyoshibhaduri@gmail.com
}

\begin{document}
\maketitle
\begin{abstract}
Known by more than 1.5 billion people in the Indian subcontinent, Indic languages present unique challenges and opportunities for natural language processing (NLP) research due to their rich cultural heritage, linguistic diversity, and complex structures. IndicMMLU-Pro is a comprehensive benchmark designed to evaluate Large Language Models (LLMs) across Indic languages, building upon the MMLU Pro (Massive Multitask Language Understanding) framework. Covering major languages such as Hindi, Bengali, Gujarati, Marathi, Kannada, Punjabi, Tamil, Telugu, and Urdu, our benchmark addresses the unique challenges and opportunities presented by the linguistic diversity of the Indian subcontinent. This benchmark encompasses a wide range of tasks in language comprehension, reasoning, and generation, meticulously crafted to capture the intricacies of Indian languages. IndicMMLU-Pro provides a standardized evaluation framework to push the research boundaries in Indic language AI, facilitating the development of more accurate, efficient, and culturally sensitive models. This paper outlines the benchmarks' design principles, task taxonomy, and data collection methodology, and presents baseline results from state-of-the-art multilingual models. As a publicly available resource, IndicMMLU-Pro\footnote{\href{https://huggingface.co/datasets/LinguaLift/IndicMMLU-Pro}{https://huggingface.co/datasets/LinguaLift/IndicMMLU-Pro}} is set to contribute significantly to advancements in Indic language-based technologies and serve as a valuable tool for the NLP community.

\end{abstract}

\begin{figure*}[t!]
    \centering
    \includegraphics[width=0.9\linewidth]{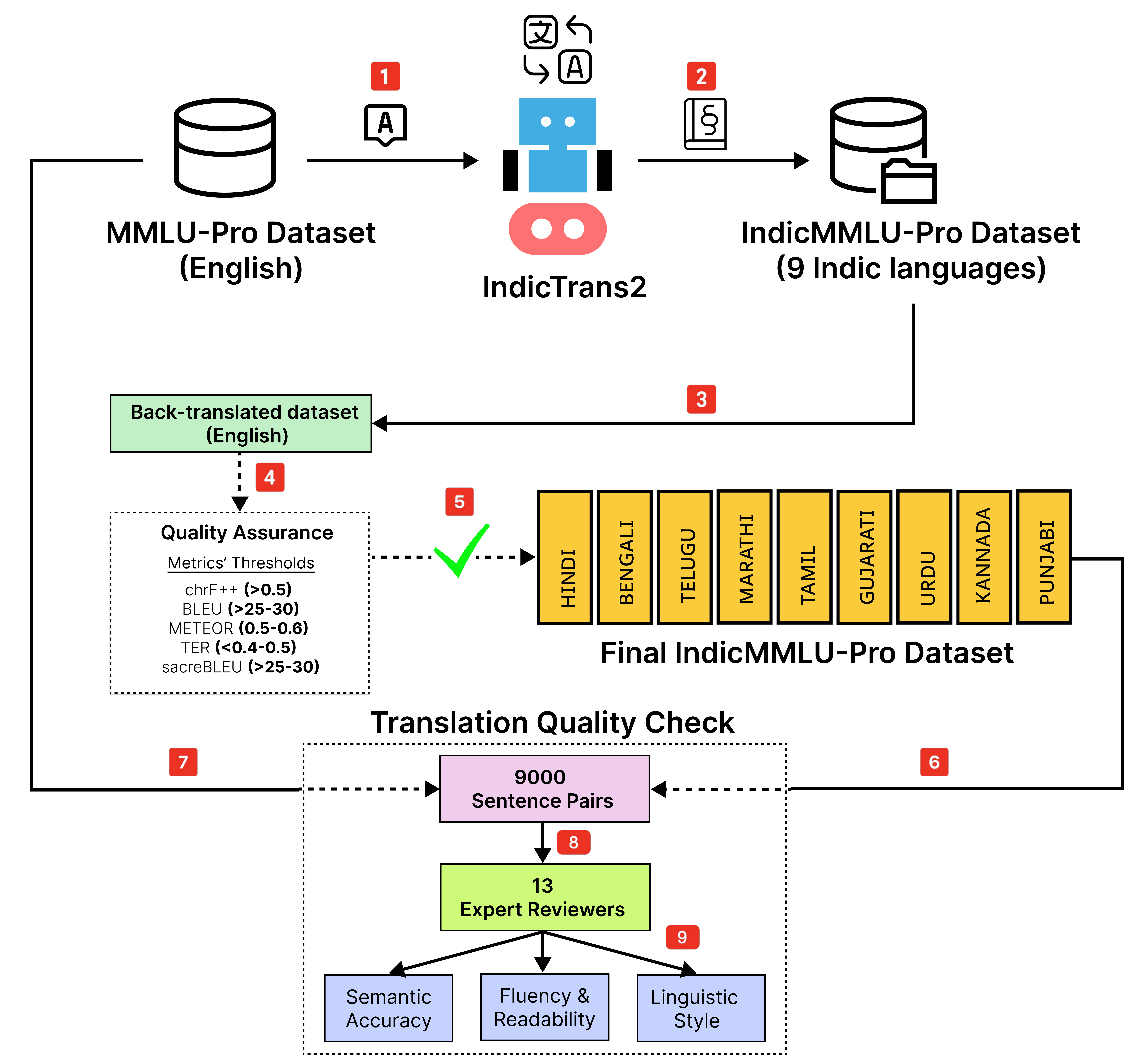}
    \caption{IndicMMLU-Pro Dataset Construction and Evaluation Pipeline. The diagram illustrates the end-to-end process of creating and validating the IndicMMLU-Pro dataset across nine Indic languages. Starting with the English MMLU-Pro dataset, content is translated using IndicTrans2 (1B parameters) and undergoes rigorous quality assurance through back-translation and multiple metric evaluations (chrF++, BLEU, METEOR, TER, and SacreBLEU). Only translations meeting quality thresholds proceed to the final dataset. The workflow also shows the comprehensive evaluation process including expert proofreading involving 13 reviewers who assess semantic accuracy, fluency, and linguistic style. This systematic approach ensures the creation of a high-quality, multilingual benchmark dataset that maintains the integrity of the original MMLU-Pro while adapting to the linguistic nuances of Indic languages.}
    \label{fig:overview}
\end{figure*}


\section{Introduction}
\label{sec:introduction}
With over 1.5 billion speakers, Indic languages constitute a substantial component of the world's linguistic tapestry, showcasing the incredible diversity of the Indian subcontinent, where languages from the Indo-Aryan and Dravidian families have evolved over centuries, shaped by a shared cultural and historical context. As an integral part of daily life, these languages facilitate communication and accessibility in various aspects of society, including education, government, media, healthcare, and social services. Further, by developing language-specific benchmarks and conducting research on Indic languages, we can increase the downstream impact on these languages on inclusion, ultimately leading to better language support and enhanced accessibility for the deaf and hard-of-hearing communities across the region who use Indic sign languages \cite{ananthanarayana2021deep}. This, in turn, can enable greater social inclusion, improved education and employment opportunities, and more effective participation in civic life for sign language users.

The disparity in Natural Language Processing (NLP) research and resources across languages is striking with a survey of existing literature showing how resources and research on Indic languages lagging \cite{kj2024decoding}. In their work, \cite{joshi2020state} revealed a stark imbalance, where a mere 28 percent of languages are considered "winners," while a staggering 88 percent are "left behind." This disparity is exemplified by the contrast between English and Bengali, languages with comparable speaker populations \cite{lane201910}. Despite this, English dominates Bengali regarding available resources, with hundreds of times more visibility on platforms like the Linguistic Data Consortium, Wikipedia, and academic publication venues of significance. 

Thus, Indic languages have historically received less attention in the field of NLP compared to more globally dominant languages \cite{das2024colonial}. The disparity in NLP resources for Indic languages has been attributed to their remarkable linguistic diversity, intricate morphology, and the scarcity of annotated datasets \cite{kakwani2020indicnlpsuite, marreddy2022resource}. Nevertheless, driven by the increasing demand for NLP applications in Indic languages and the prospect of groundbreaking technological innovations \cite{ananthanarayana2021dynamic}, there is an urgent need for rigorous evaluation benchmarks to assess the performance of AI models in this domain accurately.

To address this gap, we introduce IndicMMLU-Pro, a benchmark built upon the principles of the recently released MMLU-Pro \cite{wang2024mmluprorobustchallengingmultitask} by Tiger Labs. IndicMMLU-Pro adapts the robust multi-task principles to the unique context of Indic languages, providing a comprehensive evaluation framework that assesses the linguistic understanding, reasoning abilities, and generative capabilities of AI models.

This paper makes three key contributions. Primarily, we introduce IndicMMLU-Pro, a novel benchmark for evaluating AI models across a wide range of tasks and multiple Indic languages. Next, the design principles, task taxonomy, and data collection methodology of IndicMMLU-Pro are presented in detail, to ensure that the benchmark accurately captures the complex linguistic and cultural characteristics of Indic languages. Furthermore, we establish baseline results on IndicMMLU-Pro using state-of-the-art multilingual models, laying the groundwork for future research and development of Indic languages-based AI models.

\section{Methodology}
\label{sec:methodology}
To create IndicMMLU-Pro, a benchmark for Indic languages equivalent to MMLU Pro, we adopted a process similar to the prior work on IndicMMLU, organizing our approach into two main steps: dataset creation and baseline benchmarking.

\subsection{Dataset Creation}
\label{section:dataset_creation}
Our goal was to provide MMLU-Pro in nine Indic languages: Hindi, Bengali, Telugu, Marathi, Tamil, Gujarati, Urdu, Kannada and Punjabi. To achieve this, we used IndicTrans2 \cite{gala2023indictrans2highqualityaccessiblemachine}, a state-of-the-art machine translation model specifically designed for Indic languages.

As shown in Figure\ref{fig:overview}, we use IndicTrans2 to convert the questions and corresponding options from the original English MMLU Pro dataset into each of the target languages. This approach allowed us to maintain the structure and content of the original benchmark while adapting it to the linguistic characteristics of Indic languages.

The experimental settings for IndicTrans2 were as follows:
\begin{itemize}
    \item \textbf{Model Size}: 1B parameters
    \item \textbf{Quantization}: None
    \item \textbf{Batch Size}: 8
\end{itemize}
We applied these settings consistently across all nine language translations to ensure uniformity in the translation process.
\subsection{Quality Assurance}
To maintain the integrity and accuracy of the translated content, we implemented a rigorous quality assurance process:
\begin{itemize}
    \item \textbf{Back-translation}: For a subset of the data, we performed back-translation to English and compared it with the original text to identify any significant discrepancies.
    \item \textbf{Validation}: After back-translation, we verified the back-translated dataset with the original MMLU-Pro on numerous metrics, such as chrF++ BLEU, METEOR, TER \& SacreBLEU to ensure the dataset's quality and consistency, also described in Section \ref{sec:results}. This multi-metric evaluation provides a comprehensive assessment of the translation's accuracy and fluency.
\end{itemize}

\subsection{Dataset Structure}
The resulting IndicMMLU-Pro dataset maintains the same structure as the original MMLU-Pro dataset, with separate subsets for each of the nine Indic languages. 
The dataset retains the original categories and task types from MMLU Pro, thus ensuring identical usage and facilitating comprehensive evaluation in various domains and cognitive skills.

\subsection{Dataset Availability}
The IndicMMLU-Pro dataset is publicly available on the Hugging Face Hub \cite{kj2024indicmmlupro}. This allows for easy access and reproducibility of our results. Researchers and practitioners can directly use or adapt this dataset for their studies and applications in the processing of Indic languages.

\subsection{Baseline Benchmarking}
To establish baseline performance metrics for the IndicMMLU-Pro benchmark, we evaluated state-of-the-art multilingual language models: GPT-4o, GPT-4o-mini, Llama-3.1-8B-Instruct, IndicBERT \cite{kakwani2020indicnlpsuite}, IndicBART \cite{dabre2021indicbart}, RemBERT \cite{chung2020rethinkingembeddingcouplingpretrained}, MuRIL \cite{khanuja2021muril}, and XLM-RoBERTa \cite{conneau2020unsupervisedcrosslingualrepresentationlearning}, Navarasa, Airavata \cite{gala2024airavata}, OpenHathi, TamilLlama \cite{balachandran2023tamil}, and MahaMarathi. These models were chosen due to their demonstrated capabilities in handling multiple languages and their specific design for the inclusion of Indic languages in their training data. The benchmarking process was conducted as follows:

\begin{itemize}

\item \textbf{Model Selection}: To provide a benchmark of performance over the various languages contained in IndicMMLU-Pro, we leverage the following language models with Indic language understanding capabilities.
    \begin{itemize}
        \item \textbf{IndicBERT and IndicBART}: Specifically designed for Indic languages, offering robust performance in this domain. 
        \item \textbf{RemBERT}: A multilingual model with strong performance across diverse languages, suitable for cross-lingual tasks.
        \item \textbf{MuRIL}: A multilingual model with a focus on Indian languages, providing comprehensive coverage of Indic languages.
        \item \textbf{XLM-RoBERTa}: A large-scale multilingual model is known for its cross-lingual performance and ability to handle multiple languages efficiently.
        \item \textbf{GPT-4o and GPT-4o-mini}: State-of-the-art models with advanced capabilities in handling multilingual tasks.
        \item \textbf{Llama-3.1-8B-Instruct}: A model designed for multilingual (Hindi) instruction tasks.
        \item \textbf{Navarasa, Airavata, OpenHathi, TamilLlama, and MahaMarathi}: Models specifically designed or fine-tuned for Indic languages, enhancing their performance in this domain.
    \end{itemize}

\item \textbf{Data Preparation}: We used the \emph{test} split of the IndicMMLU-Pro dataset for each of the nine Indic languages. The data was preprocessed to match the input format required by each model.

\item \textbf{Evaluation Process}: We used accuracy as the primary metric, calculated as the percentage of correct predictions across all tasks in the benchmark. We acknowledge the importance of using multiple evaluation metrics to garner a comprehensive understanding of text data \cite{bhaduri2024reconciling}, as different metrics can capture distinct aspects of model performance \cite{bedemariam2025potential}. Nevertheless, for this study, accuracy served as a suitable baseline metric. Notably, our evaluation was conducted separately for each language, enabling language-specific performance analysis and facilitating a more nuanced understanding of model strengths and weaknesses across diverse linguistic contexts.

\item \textbf{Computational Resources}: The benchmarking was performed on a cluster of NVIDIA A100 GPUs, with each model evaluation taking approximately 24 hours per language.
\end{itemize}
The results of this baseline benchmarking are presented in Table 1 of the Results section, showing the accuracy scores for each model across all nine languages. This benchmarking process provides a solid foundation for understanding the current capabilities of multilingual models in Indic languages and sets a baseline for future research and improvements in this area.

\subsection{Proofreading and Setup}

We conducted a proofreading exercise on 9,000 sentence pairs (English and Indic), where 1,000 of these pairs were from the test split of each of the nine languages in the IndicMMLU-Pro dataset. These sentences pairs were stratified equally across the 14 categories and their respective 4 sources in each of the nine languages. These sentence pairs were shuffled randomly across all Indic languages to eliminate any bias and ensure there was no preservation for each source and category after the selection process. The three criteria that were defined for proofreading scores were:
\begin{itemize}

\item \textbf{Semantic Accuracy and Correctness}: ensuring the translation conveys the exact meaning of the original text without errors, omissions, or additions.

\item \textbf{Fluency and Readability}: check if the translation reads naturally, smoothly, and is easy to understand in the target language.

\item  \textbf{Linguistic and Stylistic Appropriateness}: ensuring the tone, style, and language fit the purpose, audience, and cultural context of the text.

\end{itemize}

A total of 13 experts, representing all 9 Indic languages, participated in the proofreading exercise. The expert distribution ensured that each language had at least one expert, with four languages having an additional expert to facilitate workload sharing. Using a standardized evaluation framework, each expert assessed sentence pairs based on three criteria, assigning scores on a 5-point scale (1 = lowest, 5 = highest). Comprehensive guidelines, accompanied by illustrative samples, were provided to all experts to ensure consistency in scoring.

The participating experts were native speakers of their respective languages, with fluency in English and formal education in their native languages. To ensure consistency, experts' scores were systematically recorded in spreadsheets, which included sentence pairs and evaluations based on the three criteria. We extend our sincere gratitude to the experts listed below under Acknowledgements, for generously volunteering their time and expertise to support this research endeavor.

\section{Results}
\label{sec:results}
Table \ref{tab:metrics} presents the evaluation metrics for the IndicMMLU-Pro dataset using back-translation techniques. The chrF++ \cite{popovic-2017-chrf} and BLEU \cite{Papineni02bleu:a} scores are provided for each of the nine Indic languages (Bengali, Gujarati, Hindi, Kannada, Marathi, Punjabi, Tamil, Telugu, and Urdu). These metrics assess the quality and accuracy of the translated content in the IndicMMLU-Pro benchmark, providing insights into the dataset's linguistic fidelity across different Indic languages.
\begin{table*}[hbt!]
    \small
    \centering
    \setlength{\tabcolsep}{3.5pt}  
    \begin{tabular*}{\textwidth}{@{\extracolsep{\fill}}lccccccc@{}}
    \toprule
       \textbf{Language} & \textbf{GPT-4o} & \textbf{GPT-4o mini} & \textbf{Llama-3.1-8B} & \textbf{IndicBART} & \textbf{IndicBERT} & \textbf{RemBERT} & \textbf{MuRIL} \\ 
    \midrule
        Hindi      & \textbf{44.80} & 32.33 & 18.61  & 11.21 & 10.78 & 11.41 & 10.87 \\
        Bengali    & \textbf{44.38} & 31.11 & N/A    & 12.52 & 10.39 & 12.00 & 9.90  \\
        Punjabi    & \textbf{40.60} & 26.25 & N/A    & 11.78 & 10.36 & 11.06 & 10.36 \\
        Marathi    & \textbf{42.20} & 27.13 & N/A    & 11.65 & 10.59 & 12.93 & 11.79 \\
        Urdu       & \textbf{44.18} & 31.13 & N/A    & 12.11 & 11.63 & 11.32 & 11.20 \\
        Gujarati   & \textbf{41.77} & 28.29 & N/A    & 12.14 & 11.06 & 12.13 & 10.79 \\
        Telugu     & \textbf{41.34} & 26.78 & N/A    & 12.05 & 11.36 & 10.20 & 9.96  \\
        Tamil      & \textbf{38.46} & 35.08 & N/A    & 11.70 & 10.96 & 10.98 & 11.00 \\
        Kannada    & \textbf{38.97} & 25.75 & N/A    & 11.51 & 11.71 & 10.87 & 10.62 \\
    \bottomrule
    \end{tabular*}
    \caption{Performance comparison of language models on the IndicMMLU-Pro benchmark across nine Indic languages, including Indo-Aryan (Hindi, Bengali, Punjabi, Marathi, Urdu, and Gujarati) and Dravidian (Telugu, Tamil, and Kannada) languages. Accuracy scores are shown as percentages. Models compared include GPT-4o, GPT-4o mini, IndicBART, IndicBERT, RemBERT, MuRIL, and Llama-3.1-8B-Instruct.}
    \label{tab:lang_perf}
\end{table*}

\begin{table*}
    \centering
    \small
    \setlength{\tabcolsep}{3.5pt}  
    \begin{tabular*}{\textwidth}{@{\extracolsep{\fill}}lcccccc@{}}
    \toprule
       \textbf{Language}  & \textbf{XLM-RoBERTa} & \textbf{Navarasa} & \textbf{Airavata} & \textbf{OpenHathi} & \textbf{TamilLlama} & \textbf{MahaMarathi} \\ 
       \toprule
        Hindi &12.33  &12.43  &11.60  &11.65  &-  &-\\
        Bengali &12.68   &12.08  & -  & - &- & -\\ 
        Punjabi &12.59 &11.95  & - & -  & - & -\\
        Marathi &12.57  &11.88  & - & - & -  &11.60\\       
        Urdu  & 12.53   &10.73  & -  & - & - &  -\\
        Gujarati &11.92    &11.53  & - & -  & - & - \\
        \midrule
        Telugu &12.62   &11.77  & - & -  &11.53 &- \\
        Tamil &12.34   &12.38  & -  & -  &11.66 & -\\ 
        Kannada &13.16  &11.88  & - & -  & - & - \\
        \bottomrule
    \end{tabular*} 
    \caption{Comparison of language model performance across Indian languages, both Indo-Aryan (i.e., Hindi, Bengali, Punjabi, Marathi, Urdu, and Gujarati) and Dravidan (i.e., Telegu, Tamil, and Kannada). Scores are shown for Llama 3.1, Navarasa, Airavata, OpenHathi, TamilLlama, and MahaMarathi models where available.}
    \label{tab:lang_perf_indic1}
\end{table*}

Table \ref{tab:lang_perf} and \ref{tab:lang_perf_indic1} present the accuracy scores (in percentages) of various pre-trained language models evaluated on the IndicMMLU-Pro benchmark. The benchmark covers nine major Indic languages: Bengali, Gujarati, Hindi, Kannada, Marathi, Punjabi, Tamil, Telugu, and Urdu. The scores reflect the models' ability to handle diverse linguistic challenges specific to Indic languages.

\subsection{Overall Performance}
\textbf{Performance Range:} The accuracy scores across all models and languages now fall within a broader range of 9.90\% to 44.80\%. This wider range reflects the significant performance differences between the newer, more advanced models (like GPT-4o) and the previously evaluated models.
 
\textbf{Model Comparison:} GPT-4o consistently outperforms all other models across all languages, achieving the highest scores ranging from 38.46\% to 44.80\%. GPT-4o mini follows as the second-best performer with scores ranging from 25.75\% to 35.08\%. Among the previously evaluated models, XLM-RoBERTa remains the top performer, consistently outperforming other models across most languages with scores ranging from 11.92\% to 13.16\%.

\textbf{Language-Model Variability:} There is now a much more pronounced variability in performance across different languages and models. This suggests that language-specific characteristics and model architectures play significant roles in performance outcomes.


\subsection{Language-wise Performance}

Breaking down the performance for each language reveals interesting patterns:
\begin{itemize}
    \item \textbf{Hindi:} GPT-4o leads with 44.80\%, followed by GPT-4o mini (32.33\%). Among other models, Navarasa (12.43\%) and XLM-RoBERTa (12.33\%) perform best. 
    \item \textbf{Bengali:} GPT-4o achieves 44.38\%, with GPT-4o mini at 31.11\%. XLM-RoBERTa (12.68\%) and Navarasa (12.08\%) lead among other models. 
    \item \textbf{Telugu:} GPT-4o scores 41.34\%, GPT-4o mini 26.78\%. XLM-RoBERTa (12.62\%) outperforms Navarasa (11.77\%) and TamilLlama (11.53\%). 
    \item \textbf{Marathi:} GPT-4o reaches 42.20\%, GPT-4o mini 27.13\%. RemBERT (12.93\%) performs best among earlier models, with Navarasa at 11.88\% and MahaMarathi at 11.60\%. 
    \item \textbf{Tamil:} GPT-4o mini shows strong performance at 35.08\%, close to GPT-4o's 38.46\%. Navarasa (12.38\%) slightly outperforms XLM-RoBERTa (12.34\%), with TamilLlama at 11.54\%. 
    \item \textbf{Gujarati:} GPT-4o scores 41.77\%, GPT-4o mini 28.29\%. IndicBART (12.14\%) and RemBERT (12.13\%) perform well among earlier models. 
    \item \textbf{Urdu:} GPT-4o achieves 44.18\%, GPT-4o mini 31.13\%. XLM-RoBERTa (12.53\%) leads among other models. 
    \item \textbf{Kannada:} GPT-4o scores 38.97\%, GPT-4o mini 25.75\%. XLM-RoBERTa significantly outperforms other models (13.16\%) among earlier models. 
    \item \textbf{Punjabi:} GPT-4o reaches 40.60\%, GPT-4o mini 26.25\%. XLM-RoBERTa (12.59\%) performs best among other models. 
\end{itemize}

\subsubsection{Cross-linguistic Analysis}
Indo-Aryan languages (Hindi, Bengali, Punjabi, Gujarati, Urdu, Marathi) and Dravidian languages (Kannada, Tamil, Telugu) exhibit distinct performance patterns.
\begin{itemize}

\item \textbf{Performance Patterns:}
GPT-4o and GPT-4o mini consistently outperform all other models across both language families.Among other models, XLM-RoBERTa generally performs best across both language families. Navarasa shows competitive performance, particularly for the Dravidian language

\item \textbf{Language-specific Models:} TamilLlama and MahaMarathi show promise but fail to surpass the performance of top multilingual models like XLM-RoBERTa and Navarasa.

\item \textbf{Consistency:} GPT-4o and GPT-4o mini demonstrate relatively consistent performance across different languages, suggesting robust multilingual capabilities. Navarasa also shows consistency, particularly among Dravidian languages.

\item \textbf{Script-based Patterns}: Languages using the Devanagari script (Hindi, Marathi) exhibit similar performance patterns, while Urdu (using the Perso-Arabic script) shows more varied performance.

\item \textbf{Performance Gap:} A significant performance gap exists between GPT-4o/GPT-4o mini and other models, highlighting the advantages of larger, more advanced models in handling the complexities of Indic languages.
\end{itemize}

\begin{table*}
    \centering
    \small
    \begin{tabular}{p{2cm}p{2cm}p{2cm}p{2cm}p{2cm}p{2cm}p{2cm}}
    \toprule
       \textbf{Language}  & \textbf{chrF++} & \textbf{BLEU} & \textbf{METEOR} & \textbf{TER} & \textbf{SacreBLEU}  \\
       \toprule
        Hindi &78.06 &0.59 &0.56 &42.27 &59.07  \\
        Gujarati &77.67 &0.58 &0.55 &43.09 &58.28 \\
        Tamil &74.32 &0.54 &0.52 &46.41 &53.64  \\
        \bottomrule
    \end{tabular}
    \caption{Back-translation evaluation metrics for the IndicMMLU-Pro dataset for 3 Indic languages.}
    \label{tab:metrics}
\end{table*}

\subsection{Dataset Quality Assessment}
In order to heuristically assert the quality of Indic language datasets, we convert the dataset from the respective Indic language back to English again by leveraging IndicTrans2 as in Section \ref{section:dataset_creation}. The English translations are then scored by the following metrics:
\begin{itemize}
    \item \textbf{chrF++} \cite{popovic-2017-chrf} \textit{(threshold > 50\%)}: A language and token agnostic, character n-gram F1 score. We use n=6, as it is the standard set in the chrF++ paper.
    \item \textbf{BLEU} \cite{Papineni02bleu:a} \textit{(threshold > 25-30\%)}: A score obtained by measuring the precision of n-grams of the candidate translation compared to the reference translation, with a brevity penalty to penalize short translations.
    \item \textbf{METEOR} \cite{banerjee2005meteor}  \textit{(threshold = 50-60\%)}: A metric that evaluates translation quality by considering precision and recall, along with a harmonic mean of unigram matches, with additional features such as stemming and synonymy matching.
    \item \textbf{TER} \cite{snover2006study}  \textit{(threshold < 40-50\%)}: A score that measures the number of edits required to change a system-generated translation to exactly match a reference translation.
    \item \textbf{SacreBLEU} \cite{post2018call} \textit{(threshold > 25-30\%)}: A variant of BLEU that ensures standard reference text processing and tokenization.
\end{itemize}

Table \ref{tab:metrics} provides insights into the quality of the IndicMMLU-Pro dataset through back-translation evaluation metrics for three languages: Hindi, Gujarati, and Tamil. The metrics include chrF++, BLEU, METEOR, TER, and SacreBLEU scores.

Hindi shows the highest overall quality with a chrF++ score of 78.06, BLEU of 0.59, METEOR of 0.56, TER of 42.27, and SacreBLEU of 59.07. Gujarati follows closely with very similar scores: chrF++ of 77.67, BLEU of 0.58, METEOR of 0.55, TER of 43.09, and SacreBLEU of 58.28. Tamil, while still demonstrating good quality, shows slightly lower scores across all metrics: chrF++ of 74.32, BLEU of 0.54, METEOR of 0.52, TER of 46.41, and SacreBLEU of 53.64.

The high chrF++ scores (above 74) for all three languages indicate good overall translation quality and semantic preservation. The BLEU and SacreBLEU scores, ranging from 0.54 to 0.59 and 53.64 to 59.07 respectively, suggest reasonably good translation quality, though there's room for improvement. The METEOR scores (0.52-0.56) also indicate good semantic similarity between the original and back-translated texts.

The Translation Error Rate (TER) scores, ranging from 42.27 to 46.41, suggest that a moderate amount of editing is required to match the reference translation, with Tamil requiring slightly more editing than Hindi or Gujarati.

It is important to note that data for the other six languages (Bengali, Punjabi, Kannada, Telugu, Urdu, and Marathi) is missing from Table 2. This limitation in the dataset quality assessment makes it challenging to draw comprehensive conclusions about the overall quality of the IndicMMLU-Pro dataset across all nine languages.

In summary, the available metrics indicate good translation quality for Hindi, Gujarati, and Tamil, with Hindi showing the highest quality across all metrics. However, the lack of data for the remaining languages highlights the need for a more comprehensive evaluation of the entire dataset.

\begin{figure*}[ht!]
    \centering
    \includegraphics[width=\linewidth]{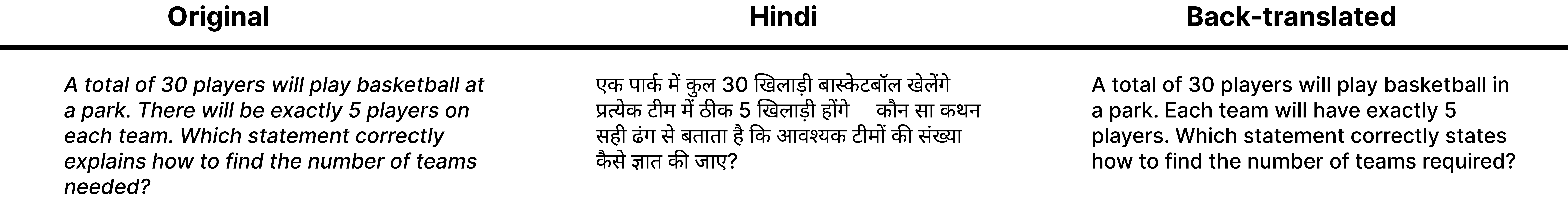}
    \caption{The original text sample, its Hindi translation, and the corresponding back-translated text}
    \label{fig:hindi_trans1}
\end{figure*}

\begin{figure*}[ht!]
    \centering
    \includegraphics[width=\linewidth]{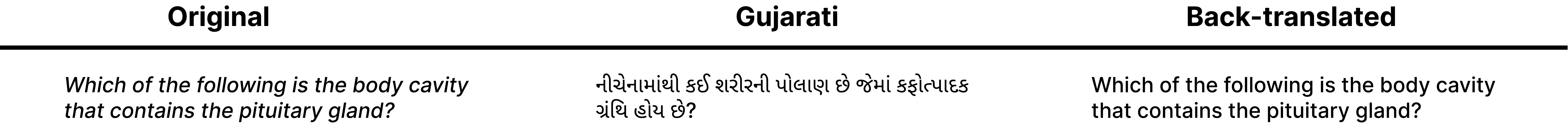}
    \caption{The original text sample, its Gujarati translation, and the corresponding back-translated text}
    \label{fig:gujarati_trans1}
\end{figure*}

\begin{figure*}[ht!]
    \centering
    \includegraphics[width=\linewidth]{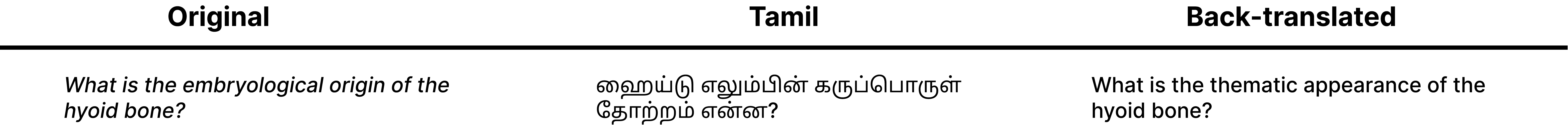}
    \caption{The original text sample, its Tamil translation, and the corresponding back-translated text}
    \label{fig:tamil_trans1}
\end{figure*}

To better illustrate the meaning of chrF++ scores in practice, let's examine sample translations for Hindi (highest score), Gujarati (middle score), and Tamil (lowest score) as shown in Figure~\ref{fig:hindi_trans1}, \ref{fig:gujarati_trans1}, \ref{fig:tamil_trans1} respectively.\\

\textbf{Hindi (chrF++: 78.06)}: The chrF++ score for Hindi is the best among the three languages indicating a high structural and semantic similarity between the original text and the back-translated text.\\

\textbf{Gujarati (chrF++: 77.67)}: While slightly lower, this score still represents a high-quality translation with minor variations.\\

\textbf{Tamil (chrF++: 74.32)}: This lower score might reflect slight changes in word choice or structure, but the core meaning is preserved.\\


These examples demonstrate that even with the lowest chrF++ score in our dataset (74.32 for Tamil), the translations maintain good semantic fidelity. The differences in scores often reflect nuances in word choice, sentence structure, or slight variations in conveying the same meaning, rather than significant errors in translation. This analysis supports our earlier observation that the IndicMMLU-Pro dataset maintains good overall translation quality across the evaluated languages.

Preliminary assessments of the other languages (Bengali, Punjabi, Kannada, Telugu, Urdu, and Marathi) suggest similar trends in translation quality, with chrF++ scores ranging from 73 to 79. Complete metrics for these languages are being compiled and will be included in future publications.

\subsection{Cosine Similarity Scores}

\begin{table}
    \centering
    \small
    \begin{tabular*}{\columnwidth}{@{\extracolsep{\fill}}lccc@{}}
       \toprule    
       \textbf{Language} & \textbf{Questions} & \textbf{Choices} \\ 
       \toprule
        Hindi &0.9109  &0.9250  \\
        Bengali &0.9172  &0.9251  \\ 
        Telugu &0.9193  &0.9287   \\
        Marathi &0.9126  &0.9242 \\
        Tamil &0.9194  &0.9255   \\ 
        Gujarati &0.9164  &0.9320 \\
        Urdu &0.9121  &0.9302   \\
        Kannada &0.9149  &0.9238  \\
        Punjabi &0.9177  &0.9254   \\
        \bottomrule
    \end{tabular*}
    \caption{Cosine similarity scores between LaBSE embeddings of IndicMMLU-Pro languages and English MMLU-Pro for questions and multiple-choice options. These scores are used as a measure of semantic similarity, with higher values suggesting closer meaning alignment across languages.}
    \label{tab:cos_sim_perf}
\end{table}

We evaluated the semantic similarity between Indic languages from the IndicMMLU-Pro dataset and the English MMLU-Pro dataset using the LaBSE (Language-agnostic BERT Sentence Embedding) model. It loads both datasets and processes all the samples from each. For each pair of corresponding questions, it generates embeddings for the questions and their multiple-choice options using the LaBSE model. 

The script then calculates cosine similarity between these embeddings to measure how semantically similar Indic languages and their english versions are. It computes two main metrics: average question similarity, and average choice similarity. The question and choice similarities indicate how close the meanings are between the two languages. This analysis helps assess how accurately Indic languages in the IndicMMLU-Pro represent the original English MMLU-Pro in terms of meaning and structure, which is crucial for ensuring the quality and consistency of multilingual datasets in machine learning and natural language processing tasks.

\section{Related Work}
\label{sec:related_work}
The study of multilingual language models has been significantly advanced by efforts to create and utilize datasets that encompass a wide range of languages. Indic languages, with their diverse scripts and rich morphological structures, present unique challenges and opportunities for multilingual NLP research.

Multilingual datasets have been crucial in developing language models capable of understanding and generating text in multiple languages. The WikiAnn dataset, for instance, provides a valuable resource for named entity recognition across many languages, including several Indic languages \cite{pan2017cross}. The IndicNLP Corpus \cite{kunchukuttan2020indicnlp}, developed as part of the IndicNLP Library, provides extensive monolingual corpora for various Indic languages. This dataset has been instrumental in training and evaluating models specifically designed for these languages. IndicCorp \cite{joshi2022indiccorp}, a large-scale dataset of Indic languages, and the IndicNLPSuite\cite{kakwani2020indicnlp}, which includes pre-trained models and linguistic resources, have further enhanced the development of multilingual NLP models. These resources have provided a solid foundation for various NLP tasks such as machine translation, text classification, etc.

IndicGLUE\cite{kakwani2020indicglue} is a comprehensive benchmark designed to evaluate the performance of NLP models on a variety of tasks across multiple Indic languages. It includes tasks such as sentiment analysis, natural language inference, and question answering, making it a valuable resource for assessing the capabilities of multilingual models.

\cite{ahuja2023megaverse} introduce Megaverse, a comprehensive benchmark designed to evaluate large language models across a variety of languages, modalities, models, and tasks. This benchmark aims to provide a holistic assessment of language models' performance, emphasizing their capabilities and limitations in handling diverse linguistic and contextual scenarios. Megaverse is instrumental in identifying the strengths and weaknesses of current models and guiding future improvements in multilingual and multimodal language processing.

\cite{holtermann2024evaluating} developed MultiQ, a benchmark aimed at evaluating the elementary multilingual capabilities of large language models. MultiQ focuses on assessing basic language understanding and generation tasks across multiple languages, providing insights into the foundational multilingual capabilities of these models. This benchmark is critical for identifying gaps in language model training and performance, particularly for less-represented languages.

\cite{aggarwal2022indicxnli} introduce IndicXNLI, a benchmark specifically designed to evaluate the natural language inference capabilities of multilingual models for Indian languages. This benchmark includes a diverse set of inference tasks, reflecting the linguistic richness and complexity of Indic languages. IndicXNLI is crucial for advancing the understanding and processing of Indian languages in NLP, providing a standard for evaluating inference models in this context.

\cite{kumar2022indicnlg} develop the IndicNLG benchmark, which provides multilingual datasets for a variety of natural language generation (NLG) tasks in Indic languages. This benchmark addresses the unique challenges posed by the linguistic diversity of Indic languages, offering datasets for tasks such as machine translation, summarization, and text generation. IndicNLG is instrumental in improving the performance and capabilities of NLG models for Indic languages, fostering advancements in multilingual NLP.

MMLU (Massively Multilingual Language Understanding) \cite{liang2020xglue, hu2020xtreme} is a comprehensive benchmark designed to evaluate the performance of language models across multiple languages and a wide range of tasks. This benchmark has been instrumental in assessing the capabilities of multilingual models and has provided insights into their strengths and limitations.

\section{Key Findings and Implications of IndicMMLU-Pro}
\label{sec:key_findings}
IndicMMLU-Pro represents a significant step forward in the development and evaluation of AI models for Indic languages. Through this comprehensive benchmark, we have shed light on the current capabilities and limitations of state-of-the-art multilingual models in handling the linguistic diversity and complexity of the Indian subcontinent. This section details implications based on some of our key findings.

\subsection{Performance Across Models}
\subsubsection{GPT-4o Dominance} GPT-4o consistently outperformed all other models across all languages, with accuracy scores ranging from 38.46\% to 44.80\%. This suggests that large, advanced language models have significant potential for handling Indic languages.

\subsubsection{Performance Tiers} A clear hierarchy emerged among the models:
\begin{itemize}
\item \textbf{Elite Tier}: GPT-4o
\item \textbf{Advanced Tier}: GPT-4o mini
\item \textbf{Specialized Tier}: XLM-RoBERTa, Navarasa, and other specialized models
\item \textbf{Foundation Tier}: Earlier models like IndicBERT, IndicBART, RemBERT, and MuRIL
\end{itemize}

\subsubsection{Specialized vs. General Models} While GPT-4o and GPT-4o mini showed superior performance, specialized models like XLM-RoBERTa and Navarasa consistently outperformed other models in the foundation tier. This highlights the importance of both scale and specialized training for Indic languages.

\subsection{Language-Specific Insights}

\subsubsection{Performance Variability} Significant variability in model performance was observed across different Indic languages, emphasizing the need for language-specific approaches in NLP.

\subsubsection{Script and Language Family Impact} Languages using similar scripts (e.g., Devanagari for Hindi and Marathi) showed similar performance patterns. Urdu, using a different script, exhibited more varied performance across models.

\subsubsection{Indo-Aryan vs. Dravidian Languages} While performance varied across both language families, some models showed more consistency in Dravidian languages, suggesting potential differences in how models handle these distinct language groups.

\subsection{Dataset and Evaluation Quality}

\subsubsection{Translation Quality} Back-translation evaluation metrics for Hindi, Gujarati, and Tamil demonstrated good overall translation quality and semantic preservation, with chrF++ scores above 74 for all three languages.

\subsubsection{Metric Discrepancies} Differences between chrF++ and BLEU scores suggest that while meaning is generally preserved, there may be variations in phrasing and structure compared to the original English text.

\subsubsection{Comprehensive Evaluation Need} The lack of evaluation metrics for six out of nine languages highlights the need for more thorough assessment across all included Indic languages.

\subsection{Methodological Insights}

\subsubsection{Benchmark Design} The adaptation of MMLU-Pro to create IndicMMLU-Pro demonstrates a successful approach to developing comprehensive, multi-task benchmarks for specific language groups.

\subsubsection{Translation Approach} The use of IndicTrans2 for dataset creation, combined with rigorous quality assurance processes, provides a template for developing high-quality multilingual datasets.

\subsubsection{Evaluation Metrics} The use of multiple metrics (chrF++, BLEU, METEOR, TER, SacreBLEU) for assessing translation quality offers a more nuanced understanding of dataset quality.

These findings and implications highlight the significant advancements made by IndicMMLU-Pro in evaluating and understanding the performance of language models across Indic languages, while also pointing to crucial areas for future research and development in multilingual NLP.

\subsection{Future Directions}

Based on our findings, we propose several key areas for future research and development:

\begin{itemize}

\item \textbf{Data Collection}: There is a pressing need for more high-quality, diverse datasets across all Indic languages, particularly for low-resource languages. This will in turn enable more robust model training and evaluation.

\item \textbf{Model Development}: Future research should focus on developing models that can better handle the unique linguistic features of Indic languages, including their complex morphology and script diversity. This may involve innovative architecture designs or novel pre-training techniques.

\item \textbf{Cross-lingual Transfer}: Exploring techniques to improve knowledge transfer between related Indic languages will help boost performance, especially for low-resource languages. This could involve leveraging linguistic similarities or developing more effective multilingual training strategies.

\item \textbf{Task-specific Fine-tuning}: Developing strategies for effective fine-tuning of large multilingual models on specific language tasks lead to significant performance improvements \cite{balne2024parameter}. Thus future work in this area may include investigating optimal fine-tuning techniques or developing Indic-specific pre-training tasks.

\item \textbf{Evaluation Metrics}: Further refinement of evaluation metrics to account for the linguistic and cultural nuances of Indic languages is necessary for more accurate performance assessment. Existing metrics often overlook the linguistic and cultural nuances inherent to Indic languages.

\end{itemize}
IndicMMLU-Pro sets a new standard for evaluating AI models in the context of Indic languages. By providing a comprehensive, multi-task, and multilingual benchmark, it enables researchers and developers to assess and improve the capabilities of their models across a wide range of Indic languages and domains. As the field of NLP continues to advance, we hope that IndicMMLU-Pro will catalyze increased research and development of technologies based on Indic languages. This benchmark not only highlights the current state of the art but also points the way forward for creating more linguistically diverse, culturally sensitive, and capable AI systems that can serve the vast and diverse population of the Indian subcontinent.

To foster a more comprehensive understanding of Indic language models, it is essential to intentionally integrate multi-disciplinary perspectives into engineering and science workstreams \cite{bhaduri2024multi}. Collaborations between engineers, scientists, and experts from social sciences can provide nuanced insights into the complex interplay between language, culture, and technology \cite{mackenzie2024beyond}. By embracing interdisciplinary approaches and refining evaluation metrics, we can work towards a more inclusive and accurate assessment of Indic language models, ultimately driving progress in natural language processing for diverse languages and communities.

In conclusion, while significant challenges remain in the field of Indic language NLP, IndicMMLU-Pro provides a robust framework for measuring progress and guiding future research efforts. As we continue to push the boundaries of what's possible in multilingual AI, benchmarks like IndicMMLU-Pro will play a crucial role in ensuring that technological advancements benefit all language communities, fostering greater inclusivity and accessibility in the digital age.

\section{Acknowledgements}
\label{sec:acknowledgements}
We extend our deepest gratitude to the experts who volunteered their time and effort to evaluate our dataset. Unfortunately, Indic languages are often neglected in the development and evaluation of Large Language Models, making initiatives like ours reliant on the generosity and expertise of individuals like those listed below. The list of experts and the language they reviewed are:

\begin{itemize}
    \item Telugu Experts: Dushyanth Reddy and Sai Krishna Bolla
    \item Tamil Expert: Veeramanohar Avudaiappan
    \item Hindi Expert: Ojasmitha Pedirappagari
    \item Gujarati Expert: Ashwini Thosar
    \item Punjabi Expert: Shivan Bhatia
    \item Marathi Experts: Sameer Jadhav and Nityanand Bhadra
    \item Kannada Expert: Nikhil Ragunath
    \item Bengali Experts: Sounak Datta and Nirjhar Barua 
    \item Urdu Experts: Md Mahtab and Risha Fatema
\end{itemize}

Expert feedback has significantly enhanced the quality of our dataset, and we are grateful for their dedication to promoting Indic languages in the global LLM community.

\bibliography{acl_latex}

\section{Appendix}

\subsection{Choice of Models}
The selection of models for our benchmark evaluation was based on their relevance to Indic languages, their prominence in multilingual NLP research, and their potential to handle the diverse linguistic features of Indic languages. Here's a detailed explanation of our choices:

\begin{itemize}
\item \textbf{GPT-4o}: A highly advanced multimodal model that offers improved speed and cost-effectiveness compared to GPT-4 Turbo, while also featuring enhanced vision capabilities. With its 128K context window and knowledge cutoff in October 2023, GPT-4o represents the cutting edge in large language models, making it an excellent candidate for benchmarking across diverse Indic languages.

\item \textbf{GPT-4o-mini}: Designed as a more cost-efficient alternative to larger models, GPT-4o-mini offers a balance between performance and resource utilization. Despite its smaller size, it outperforms GPT-3.5 Turbo in intelligence while maintaining lower costs. Its inclusion in our benchmark allows us to assess the trade-offs between model size, performance, and efficiency in processing Indic languages.

\item \textbf{IndicBERT}: A transformer-based multilingual model specifically pre-trained on 12 major Indian languages. Developed by AI4Bharat, IndicBERT is designed to capture the nuances of Indic languages through its training on a large corpus of Indian language text. Its inclusion in our benchmark is crucial for evaluating performance on tasks that require a deep understanding of Indian language structures and semantics.

\item \textbf{IndicBART}: A sequence-to-sequence model tailored for Indian languages, IndicBART extends the capabilities of mBART by incorporating additional pre-training on Indic language corpora. It excels in generation tasks such as summarization, translation, and text completion across multiple Indian languages. Its inclusion allows us to assess performance on more complex, generative NLP tasks specific to Indic languages.

\item \textbf{RemBERT}: Developed by Google Research, RemBERT (Rebalanced Multilingual BERT) is an advanced multilingual model that addresses the limitations of previous multilingual models. It uses a novel rebalancing technique during pre-training to improve performance across a wide range of languages, including low-resource ones. RemBERT's inclusion in our benchmark provides a strong baseline for assessing how well general multilingual models perform on Indic language tasks compared to Indic-specific models.

\item \textbf{MuRIL}: (Multilingual Representations for Indian Languages): Developed by Google Research India, MuRIL is specifically designed to address the unique challenges of Indian languages. It's trained on significantly larger amounts of Indian language data compared to general multilingual models, covering 17 Indian languages including low-resource ones. MuRIL's architecture incorporates techniques to handle the linguistic diversity and script complexity of Indian languages, making it a crucial benchmark for evaluating specialized Indic language performance.

\item \textbf{XLM-RoBERTa}: A large-scale multilingual model trained on 100 languages, XLM-RoBERTa represents a significant advancement in cross-lingual language understanding. Developed by Facebook AI, it uses a self-supervised learning approach on a vast amount of web-crawled data. Its inclusion in our benchmark allows us to assess how well a general multilingual model, trained on a diverse set of languages including Indic ones, performs on specific Indic language tasks compared to models with a more focused Indic language training.

\item \textbf{OpenHathi}: An open-source large language model focused primarily on Hindi and other Indian languages. Developed as part of an initiative to create freely accessible AI resources for Indic languages, OpenHathi is trained on a diverse corpus of Indian language texts, including literature, news, and web content. Its inclusion in our benchmark allows us to evaluate the performance of community-driven, open-source efforts in Indic language modeling, particularly for Hindi, and compare it with proprietary and general multilingual models.

\item \textbf{Airavata}: Developed by AI4Bharat, Airavata is a specialized large language model focused on Hindi and other Indian languages. It represents a significant effort in creating instruction-tuned models specifically for Indic languages. Trained on a diverse corpus of Hindi text and fine-tuned on carefully curated instruction datasets, Airavata aims to understand and generate contextually appropriate responses in Hindi. Its inclusion in our benchmark allows us to evaluate the effectiveness of language-specific instruction tuning for Indic languages, particularly in tasks requiring a nuanced understanding of cultural and linguistic contexts unique to Hindi.

\item \textbf{TamilLlama}: A pioneering language model specifically designed for Tamil, one of the classical languages of India. Developed by a team of researchers and language enthusiasts, TamilLlama is built upon the LLaMA architecture but fine-tuned extensively on a large corpus of Tamil text. It incorporates unique features to handle the agglutinative nature of Tamil and its complex morphology. By including TamilLlama in our benchmark, we aim to assess the advantages of highly specialized, language-specific models in capturing the intricacies of individual Indic languages, particularly for tasks that require a deep understanding of Tamil linguistic structures and cultural nuances.

\end{itemize}

\subsection{Model Output \& Evaluation}
\begin{figure*}
    \centering
    \includegraphics[width=0.9\linewidth]{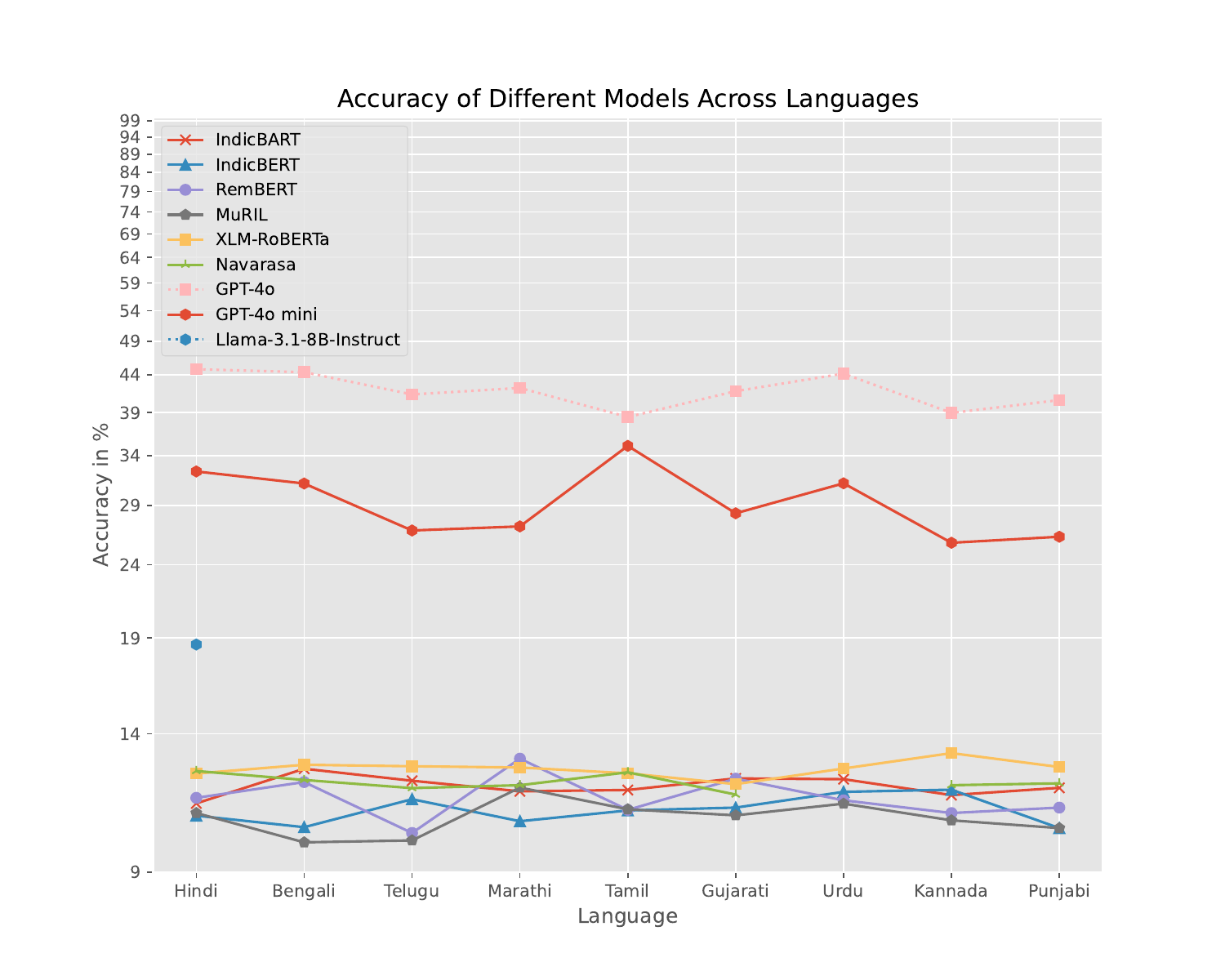}
    \caption{Model Accuracy Across Different Languages}
    \label{fig:model_perf}
\end{figure*}
\begin{figure}[ht!]
    \centering
    \includegraphics[width=\linewidth]{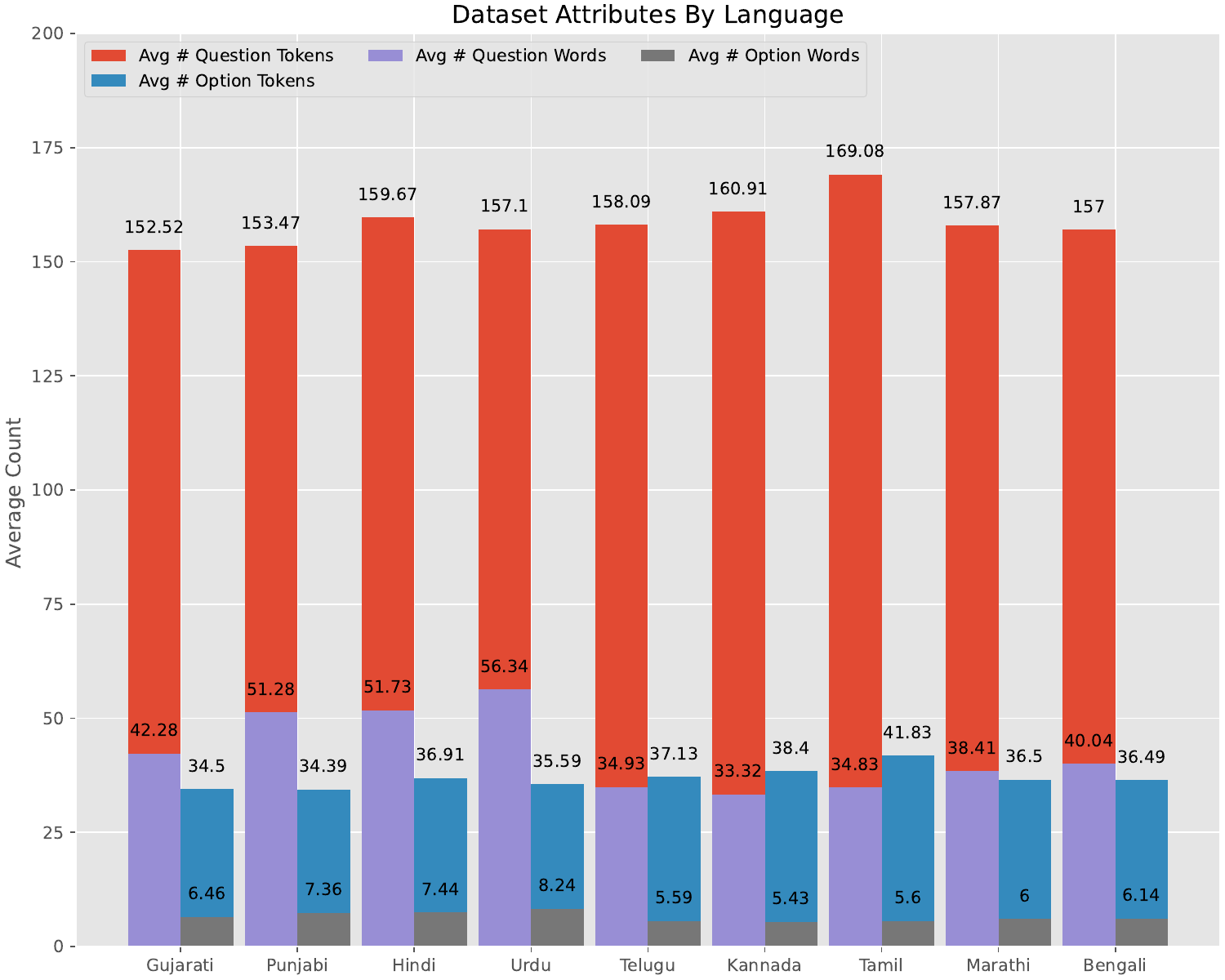}
    \caption{Average Question and Option Lengths in \# Words and \# Tokens}
    \label{fig:data_attrib}
\end{figure}
\begin{itemize}
    \item \textbf{IndicBERT:} The IndicBERT model, when used with AutoModelForMultipleChoice, outputs logits for each option in a multiple-choice question. These logits represent the model's confidence scores for each option. The model produces logits for each option. torch.argmax(logits, dim=1) selects the index of the option with the highest logit (i.e., the model's predicted answer). This index is converted to a letter (A, B, C, D) using the index\_to\_letter dictionary. The predicted answer (letter) is directly compared with the correct answer.

    \item \textbf{IndicBART:} The IndicBART model is being loaded with AutoModelForSeq2SeqLM framework to construct input texts by combining the question with each of its options in the format: 'Question: <question> Answer: <option>'. These input texts are encoded into a format that the model can process. The model generates responses for each encoded input text, effectively simulating an answer for each option. A scoring heuristic, such as the length of the generated response, is then applied to determine the relevance or quality of each simulated answer. The option with the highest score is selected as the model's predicted answer. This method leverages the model's ability to generate coherent and contextually appropriate responses, allowing it to adapt to different questions and options effectively. It iterates through all questions, applies this process, and calculates the accuracy by comparing predicted answers with the correct ones, thus evaluating the model's performance in handling multiple-choice questions across various languages.

    \item \textbf{MURIL:} The MURIL (Multilingual Representations for Indian Languages) model is being used for sequence classification. The model's output is a single logit value for each question-option pair, representing the model's confidence or relevance score for that pair. The evaluation process works as follows: For each question, the script passes the question text paired with each option to the model separately. The model produces a score for each pair. These scores are collected for all options of a question. The option with the highest score is selected as the predicted answer. This approach assumes that the option most relevant or coherent with the question (as determined by the model's score) is likely to be the correct answer. While this method can work for some types of questions, it's not the standard way to use MURIL for multiple-choice questions. MURIL is primarily designed for tasks like text classification, not specifically for multiple-choice question answering. A more appropriate approach would be to fine-tune MURIL on a multiple-choice QA task, or to use a model specifically designed for this purpose. The current method might work to some extent due to MURIL's pretraining on diverse tasks, but it may not be as accurate or reliable as a properly fine-tuned model for multiple-choice questions.

    \item \textbf{RemBERT:} The RemBERT model, which is designed for multiple-choice tasks, is being used. The model's output is a set of logits, one for each option in the multiple-choice question. These logits represent the model's confidence scores for each option. The evaluation process works as follows: For each question, the script encodes the question paired with each option and passes this to the model. The model then outputs logits for all options simultaneously. The option with the highest logit score is selected as the predicted answer. This approach is correct and appropriate for multiple-choice question-answering tasks, as it allows the model to consider all options together in context, rather than evaluating them independently. The model has been trained to understand the relationship between the question, the correct answer, and the distractors, making it well-suited for this type of task. This method of using RemBERT for multiple-choice questions is a standard and effective approach in natural language processing for question-answering tasks.

    \item \textbf{XLM-RoBERTa:} The XLM-RoBERTa model is being used for sequence classification. The model's output is a single logit value for each question-option pair, representing the model's relevance or compatibility score for that pair. The evaluation process works as follows: For each question, the script passes the question text paired with each option to the model separately. The model produces a score for each pair. These scores are collected for all options of a question. The option with the highest score is selected as the predicted answer. This approach assumes that the option most relevant or coherent with the question (as determined by the model's score) is likely to be the correct answer. While this method can work to some extent, it's not the optimal way to use XLM-RoBERTa for multiple-choice questions. XLM-RoBERTa is primarily designed for tasks like text classification or sequence pair classification, not specifically for multiple-choice question answering. A more appropriate approach would be to fine-tune XLM-RoBERTa on a multiple-choice QA task, or to use a model architecture specifically designed for multiple-choice tasks, which would consider all options simultaneously rather than in pairs.

    \item \textbf{Airavata} We evaluate the performance of the Airavata language model on the Hindi subset of the IndicMMLU-Pro dataset. It loads a specified model using the Hugging Face transformers library, and then processes multiple-choice questions from the dataset. For each question, it creates a prompt in a chat format, feeds it to the model, and compares the model's single-token prediction to the correct answer. The script calculates the overall accuracy of the model's predictions across all questions in the dataset. Finally, it saves this accuracy score to a text file and prints it to the console, providing a quantitative measure of the model's capability in understanding and answering Hindi multiple-choice questions.

\end{itemize}
\section{Additional Data Samples}
\begin{figure*}[ht!]
    \centering
    \includegraphics[width=\linewidth]{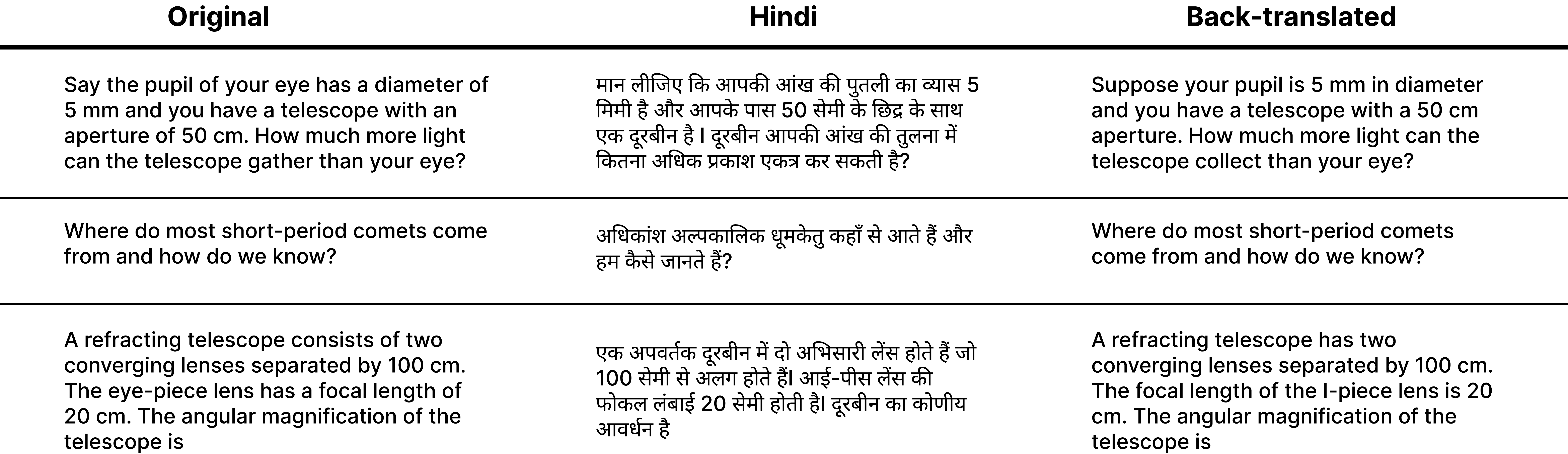}
    \caption{Additional examples showcasing the machine translation workflow, including the original text samples, their Hindi translations, and the corresponding back-translated texts.}
    \label{fig:hindi_trans}
\end{figure*}

\begin{figure*}[ht!]
    \centering
    \includegraphics[width=\linewidth]{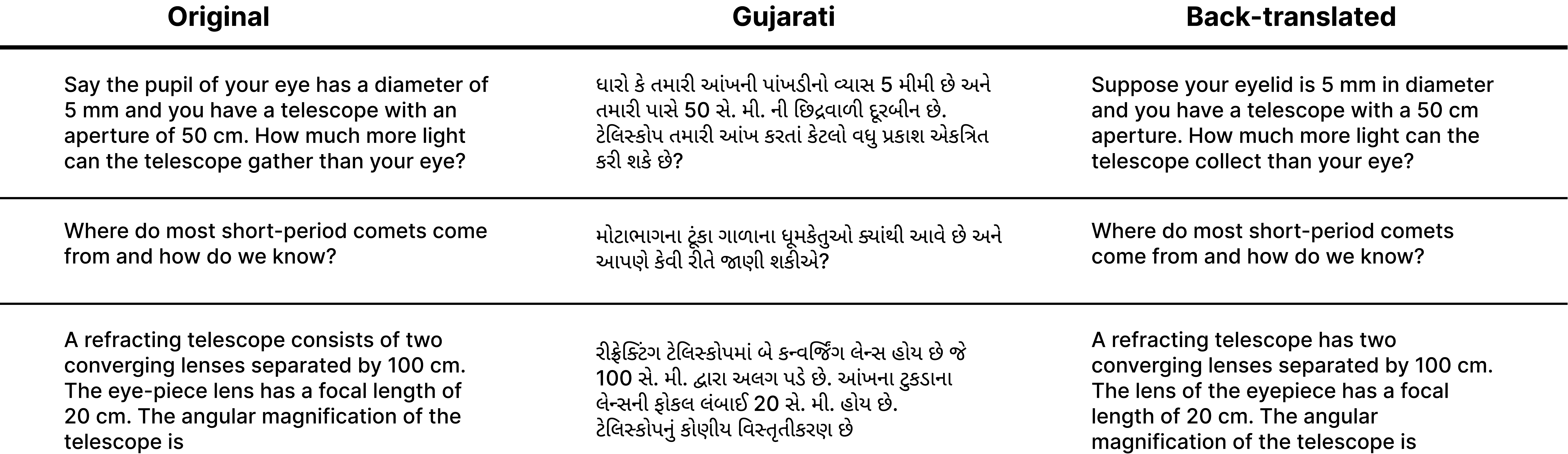}
    \caption{Additional examples showcasing the machine translation workflow, including the original text samples, their Gujarati translations, and the corresponding back-translated texts.}
    \label{fig:gujarati_trans}
\end{figure*}

\begin{figure*}[ht!]
    \centering
    \includegraphics[width=\linewidth]{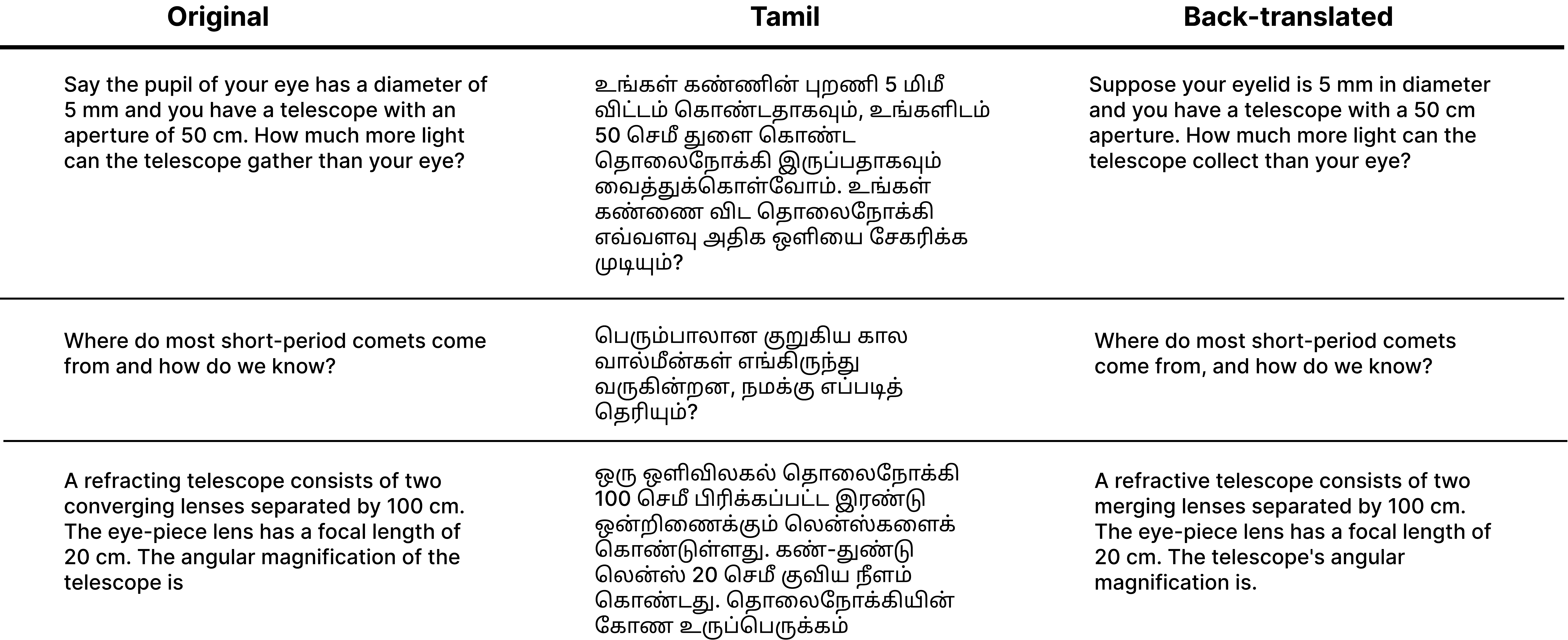}
    \caption{Additional examples showcasing the machine translation workflow, including the original text samples, their Tamil translations, and the corresponding back-translated texts.}
    \label{fig:tamil_trans}
\end{figure*}

To further illustrate the performance of the machine translation workflow, we provide examples of translations and back-translations in three different languages: Hindi, Gujarati, and Tamil. Figure \ref{fig:hindi_trans1} showcases the original text sample, its translation into Hindi, and the corresponding back-translated text, demonstrating the transformation and consistency of the translation process. For additional insights, Figure \ref{fig:hindi_trans}, \ref{fig:gujarati_trans}, \ref{fig:tamil_trans} includes a broader set of examples across Hindi, Gujarati, and Tamil, highlighting the system's ability to handle linguistic diversity and maintain semantic fidelity across these languages. These examples emphasize the efficacy and robustness of the translation pipeline in preserving the original meaning during both forward and reverse translation.



\end{document}